\documentclass[conference, letterpaper]{IEEEtran}
\usepackage{cite}

\ifCLASSINFOpdf
\else
  \usepackage[dvips]{graphicx}
  \graphicspath{{resource/}}
  \DeclareGraphicsExtensions{.eps}
\fi
\usepackage[cmex10]{amsmath}
\usepackage[tight,footnotesize]{subfigure}
\ifCLASSINFOpdf
   \usepackage[pdftex]{graphicx}
\else
\fi

\usepackage[cmex10]{amsmath}
\usepackage{color}
\usepackage{fancyhdr}

\renewcommand{\thispagestyle}[2]{}

\fancypagestyle{plain}{
        \fancyhead{}
        \fancyhead[C]{first page center header}
        \fancyfoot{}
        \fancyfoot[C]{first page center footer}
}
\usepackage{soul}
           
\begin{document}

%
\title{Measuring Player's Behaviour Change over Time in Public Goods Game}


\author{\IEEEauthorblockN{Polla Fattah}
\IEEEauthorblockA{School of Computer Science\\
The University of Nottingham\\
Nottingham, UK \\
psxpf1@nottingham.ac.uk}
\and
\IEEEauthorblockN{Uwe Aickelin}
\IEEEauthorblockA{School of Computer Science\\
The University of Nottingham Ningbo’\\
China, Ningbo \\
uwe.aickelin@nottingham.edu.cn}
\and
\IEEEauthorblockN{Christian Wagner}
\IEEEauthorblockA{School of Computer Science\\
The University of Nottingham\\
Nottingham, UK \\
pszcw@nottingham.ac.uk}}


%


\maketitle

\begin{abstract}
An important issue in public goods game is whether player's behaviour changes over time, and if so, how significant it is. In this game players can be classified into different groups according to the level of their participation in the public good. This problem can be considered as a concept drift problem by asking the amount of change that happens to the clusters of players over a sequence of game rounds. In this study we present a method for measuring changes in clusters with the same items over discrete time points using external clustering validation indices and area under the curve. External clustering indices were originally used to measure the difference between suggested clusters in terms of clustering algorithms and ground truth labels for items provided by experts. Instead of different cluster label comparison, we use these indices to compare between clusters of any two consecutive time points or between the first time point and the remaining time points to measure the difference between clusters through time points. In theory, any external clustering indices can be used to measure changes for any traditional (non-temporal) clustering algorithm, due to the fact that any time point alone is not carrying any temporal information. For the public goods game, our results indicate that the players are changing over time but the change is smooth and relatively constant between any two time points.
\end{abstract}


\begin{IEEEkeywords}
 clustering; external cluster validity; measuring change over time; temporal data
\end{IEEEkeywords}

%
\IEEEpeerreviewmaketitle

\section{Introduction}
In experimental economics there is an interest in how players of public goods game change their strategy during multiple rounds of the game play and jump from one type of player into another \cite{Fischbacher2010}, such as changing from conditional co-operator to free rider behaviour (game and player types are described in detail in later sections). This change can be seen as a drift from the original label assigned to the players. As defined by Widmer et al. \cite{Widmer1996}, concept drift is an unexpected change from the targeted future estimation due to uncalculated hidden contexts in the system. Tsymbal \cite{Tsymbal2004} identified two types of concept drift: sudden and gradual. This work presents a method to measure the quantity of the change occurring within populations in any two different time points.
 
There are many methods in machine learning for classification, with the existence of concept drift \cite{Elwell2011,Garnett2008,Xiaofeng2014} and methods to detect it \cite{Baena-Garcia2006,Harel2014}. Moreover measuring changes in clusters for different time points are well studied in data analysis, especially for data streams \cite{Ntoutsi2011,Spiliopoulou2013,Yang2011}. However, these methods aim to find overall patterns of change in clusters' location, size, merging, emerging and/or dissipating rather than presenting a measure of how much change has occurred in each cluster (i.e. in which ratio items change their membership from one cluster into another).

External cluster validity is primarily used to check the performance of clustering algorithms by measuring the difference between ground truth labels given to the items by experts and the group in which they have been placed by a clustering algorithm \cite{Halkidi2002a}. This study uses external cluster validity measures like variation of information \cite{Meila2003}and area under the curve of receiver operating characteristic \cite{Bradley1997} as scaler measures, to show the amount of items that jumped from one cluster to another between two consequent time points. To accomplish this measurement, the items should be clustered separately in every time point. As the clustering is performed at a single time point, which eliminates the time dimension for the collected data about items, theoretically any traditional (non-temporal) clustering algorithm should be sufficient. After clustering, an external clustering validity measure can quantify the amount of changes between clusters at any two time points.
With the public goods game data, each game round is used as a time point so that players are clustered in each time point using k-means algorithm, then clusters of each time point are compared with the first round to measure the amount of change in the players' strategies using multiple external clustering validity measures and area under the curve. The results show that players' strategies (their original clusters in the first round) change from one time point into another at a slow rate.
To compare our results, public goods games data were also tested using MONIC, which is a method of detecting changes among clusters in the data stream. The results show that there is a periodic change in clusters as they disappear and other clusters are emerging, but this is inconclusive as there is no indication of whether the change originated from players' strategies or from the nature of algorithm, as it reduces the effect of the old items in the cluster and removes them after two time points.

\section{Background and Related Work}

This work uses data of public goods game as a case study and multiple data mining principles like external cluster validity, area under the curve and data analysis methods of detecting changes of clusters in the data streams. These subjects are reviewed in the next few sections, providing the relationship between the subject and our study.

\subsection{Public Goods Game}
Public good is any type of services or resources that cannot be withheld from individuals without competition to benefit from these resources and services due to their characteristics of being non-rivalry and non-excludable \cite{Kaul1999}. Examples for public goods are: city parks as all citizens can attend the park while a fraction of them are paying to maintain them, and Street Lights which is useful for everybody while only tax payers are contributing in keeping them alive. Public goods game is an experimental game that simulates real situations of public good in a lab with restricted conditions and focused purposes to conduct experiments. There are many slightly different variations of this game, however the data which has been used in this paper as a case study is based on the model of Fischbacher et al \cite{Fischbacher2012}.

The public goods game experiment of Fischbacher et al \cite{Fischbacher2012} consists of four players, each of whom has a choice to contribute in a project representing the public good. After all players made their choices of contribution the game will finish and their final outcome will be revealed to them. Players are then redistributed to play with other new partners for another round of the game. However, players might adjust their strategy of contribution and learn general players' behaviour in previous games. For every round each player has 20 tokens to play with representing money, which they can contribute with, and after the end of the experiment they will be exchanged with a rate into real money to ensure that players are playing thoughtfully.

Gaining the maximum amount of tokens is the main goal of each player, and it is the basis for determining whether players change their behaviour in the next round or not. As each player has 20 tokens, they can contribute all, none or any amount to projects representing public good, so that the total amount of contribution of all players and its extra benefit will be distributed between them evenly. The amount of gain for a player i ($gain_i$) is demonstrated by the equation $gain_i=20 - g_i + 0.4\sum_{j=1}^{4} g_j$ , where $g_i$ is the player's own contribution and $g_j$ represents all players' contributions. To illustrate this equation: (1) if no player contributes in the project then each will end up with 20 tokens as they started; (2) if all players contribute with 10 tokens then each player will end up with {20-10+0.4 (10+10+10+10)} = 26 tokens; and (3) if only one player contributes with all 20 tokens while the others do not contribute, then she will end up with 8 tokens while all others will gain 28 tokens.

However, regardless of players' potential adjustment of their contribution behaviour during multiple rounds (10 rounds or more), economists \cite{Fischbacher2001} classify them based on a contribution table of static data filled once by the players prior to the game rounds. This table consists of players' answers for a hypothetical rounded average contribution of others. That is, for each possible contribution from 0 to 20 tokens, as an average, from her partners she should decide how much she is willing to contribute. Naturally, this initial willingness for contribution might change due to the factor of learning about other players' contribution behaviour, which causes concept drift throughout game time points (rounds).  The classes as defined by economists are:

\begin{itemize}
\item Conditional Co-operator: players who show more willingness to contribute when other players contribute more.
\item Free Riders: players who do not contribute to the project regardless of other players' contribution status.
\item Triangle Contributors: players whose contribution rise to a point then starts to decline afterward regarding other players' contribution.
\item Others: players with no clear pattern in their contribution style.
\end{itemize}

\subsection{External Cluster Validity}

External criteria validate results based on some predefined structure for data that is provided beside clustered data in form of labelling. The main task of this approach is to determine a statistical measure for the similarity or dissimilarity between obtained clusters and labels \cite{Halkidi2002a, Rendon2011}.

\subsubsection{Variation of Information}
This index measure is based on contingency table which is a matrix with $r \times k$, where r is number of produced clusters and k is number of externally provided clusters. Each element of this matrix contains number of agreed instances between any two clusters of the externally provided and produced clusters. As introduced by Meilă [19], this index calculates mutual information and entropy between previously provided and produced clusters derived from contingency table $VI(C,T)= 2H(T,C)- H(T)- H(C)$, where $C$ is produced clusters and $T$ is ground truth clusters, $H(C)$ is entropy of $C$ and $H(T)$ is entropy of $T$.

\subsubsection{Pair-Wise Measures}
There are multiple external clustering validity measures (e.g. Jaccard Coefficient, Fowlkes-Mallows Measure and Rand Statistic) which use the partition and cluster label information over all pairs of points \cite{Zaki2014}. If any two pairs of points are in the same cluster and have the same label then they will be counted as true positive (TP). If they are in the same cluster but have different labels, they are false positive (FP). If a pair of points is in different clusters and each point has different labels, this is true negative (TN); otherwise, the pair is false negative (FN) \cite{Vendramin2010}. Each of these measures uses the following equations:

\begin{itemize}
\item Jaccard Coefficient is function of TP pairs over all pairs except for TN So that $JC = \dfrac{TP}{TP+FN+FP}$.
\item Rand Statistic is fraction of agreed pairs (that is TP and TN) over all pairs. So that $RS = \dfrac{TP+TN}{N}$ where $N$ is number all pairs.
\item Fowlkes-Mallows Measure is a function of overall precision and recall values, which makes $FM = \sqrt{recall \times pricition}$ where $recall=  \dfrac{TP}{TP+FP}$ and $pricition=  \dfrac{TP}{TP+FN}$ so that $MF=  \dfrac{TP}{\sqrt{(TP + FN)(TP +FP)}}$ .

\end{itemize}

\subsubsection{Area Under the Curve}
Area Under the Curve (AUC) of receiver operating characteristic (ROC) is a scalar measure originally used by Bradley \cite{Bradley1997} to calculate the performance of machine learning algorithms, such as classification. The ROC curve is a graph of true positive rate (TPR) and false positive rate (FPR) of the predicted classifier's result compared to the real class for each item, so that AUC is the area under the ROC curve whose value can be between 0 and 1. Methods of calculating AUC vary according to the nature of application and available data. In this study we implemented R language \cite{Robin2014} using the trapezoidal method of Fawcett \cite{Fawcett2006}. The multi-class AUCs are calculated using the equation of Hand and Till \cite{Hand2001}  $auc= \dfrac{c}{c(c-1)} \sum aucs$, where $c$ is number of clusters and $aucs$ is a set of AUC between any two classes.

\subsubsection{Measuring Cluster Changes in Data Streams}

Many techniques and methods are introduced to track cluster changes in data streams, each of which focuses on different aspects of change that happen in clusters, like location, dimension and existence. Furthermore, different ways of dealing with data and clustering have been presented, such as MONIC model \cite{Spiliopoulou2006}, which targets detecting cluster transition over accumulated data. This method provides an ageing function for clustering data, which prioritizes new records over old ones, and eliminates records older than two time points. This model relies on accumulated data over time to detect cluster matches; therefore, it cannot be used with non-accumulated data. Moreover it emphases on measuring cluster changes and it cannot detect changes in cluster membership for individual items which have been clustered over time points.

Another method introduced by Günnemann et al \cite{Gunnemann2011a} traces cluster evolution as change in behaviour of items' recorded values in high dimensional data sets. Instead of object identifier between consequent snapshots of data, different types of mapping functions are used to map clusters according to their values in different dimensions and subspaces. This method counts number of various changes that occur to clusters of any high dimensional data set and it has an advantage of functioning even for items without having to track between snapshots, but it lacks any mean of quantifying the changes themselves. In other words, there is no indication for the quantity of change that happens to any cluster in two consecutive time points.

Hawwash and Nasraoui \cite{Hawwash2012a} introduced a framework of data stream mining using statistical cluster measures like cardinality, scale and density of clusters to detect milestones of clusters change and monitoring the behaviour of clusters. It targets accumulative clustering on data streams, but instead of using fixed time window for clustering, it uses milestones to detect next best clustering time.

Another method presented by Kalnis et al \cite{Kalnis2005} for clustering moving objects in the snapshots of spatio-temporal data is cluster mapping function, by which clusters are treated as sets and the cardinality ratio of intersection is calculated at each two time constitutive clusters over their union. If the ratio passes a certain threshold, the cluster is considered to be a moving cluster. This method detects ''move'' in overall clusters with visual aids, which assists human experts in grasping changes in the underlying data, which is good for tracking one type of cluster change for moving clusters. However, it does not provide mean to quantify the magnitude of change for overall clustering objects.

The previous methods detect changes in temporal data by monitoring cluster changes as a whole in terms of split, absorbed, disappear, emerged etc., which is a good indication for detecting change, but it does not specify its magnitude.
\begin{figure*}[t]
\centering
\subfigure[First time point]{%
\includegraphics[width=0.32\textwidth]{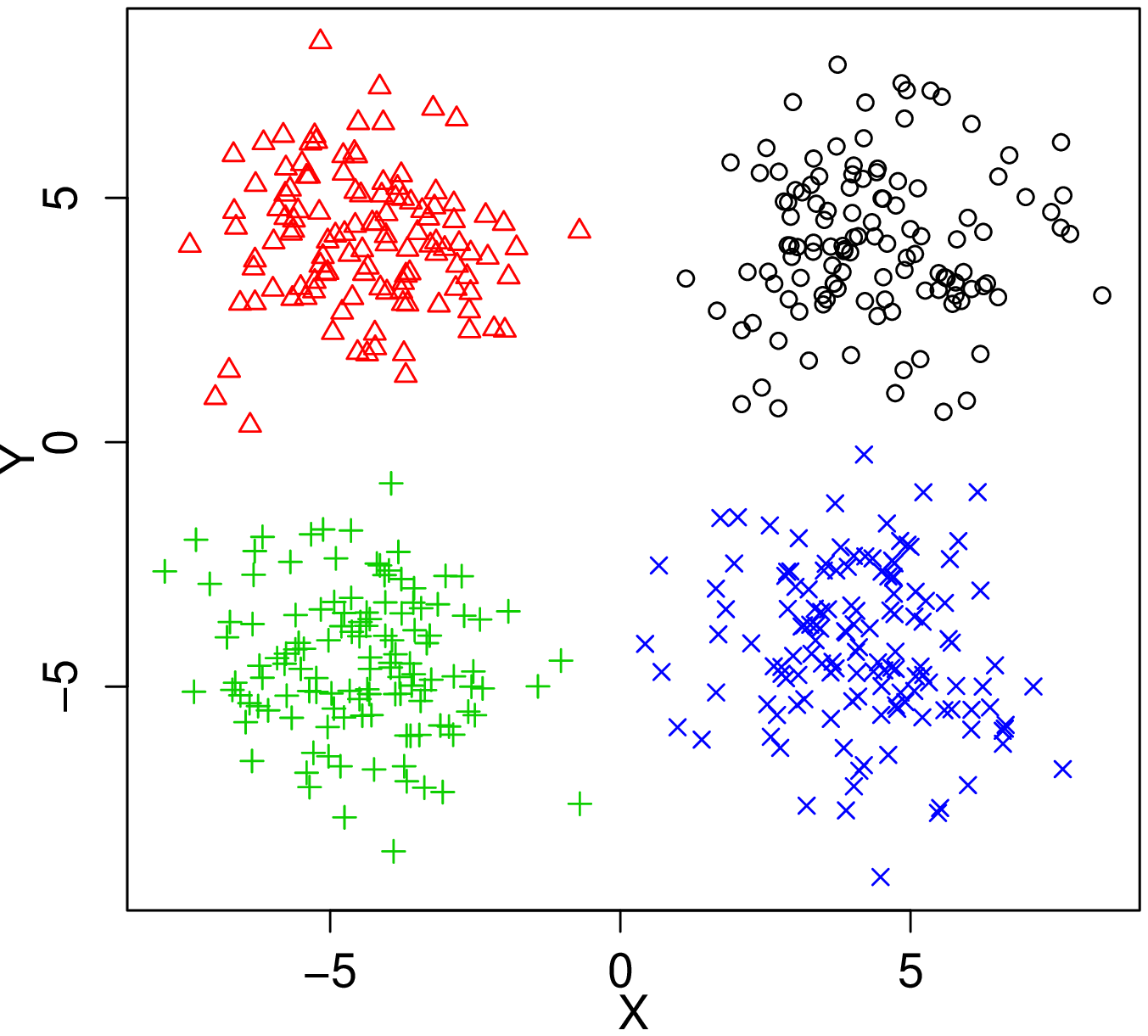}}
\subfigure[Middle time point]{%
\includegraphics[width=0.32\textwidth]{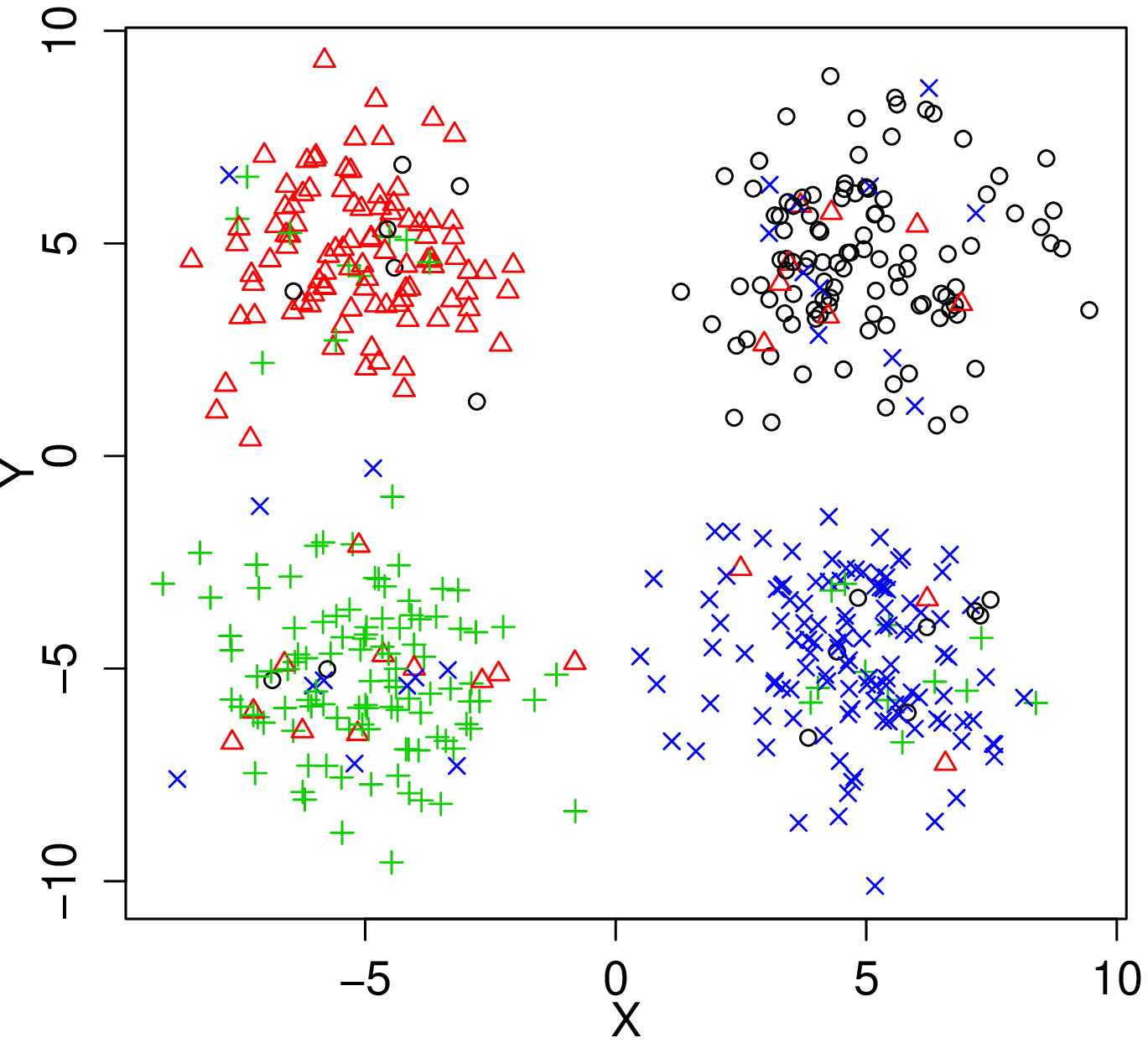}}
\subfigure[Last time point]{%
\includegraphics[width=0.32\textwidth]{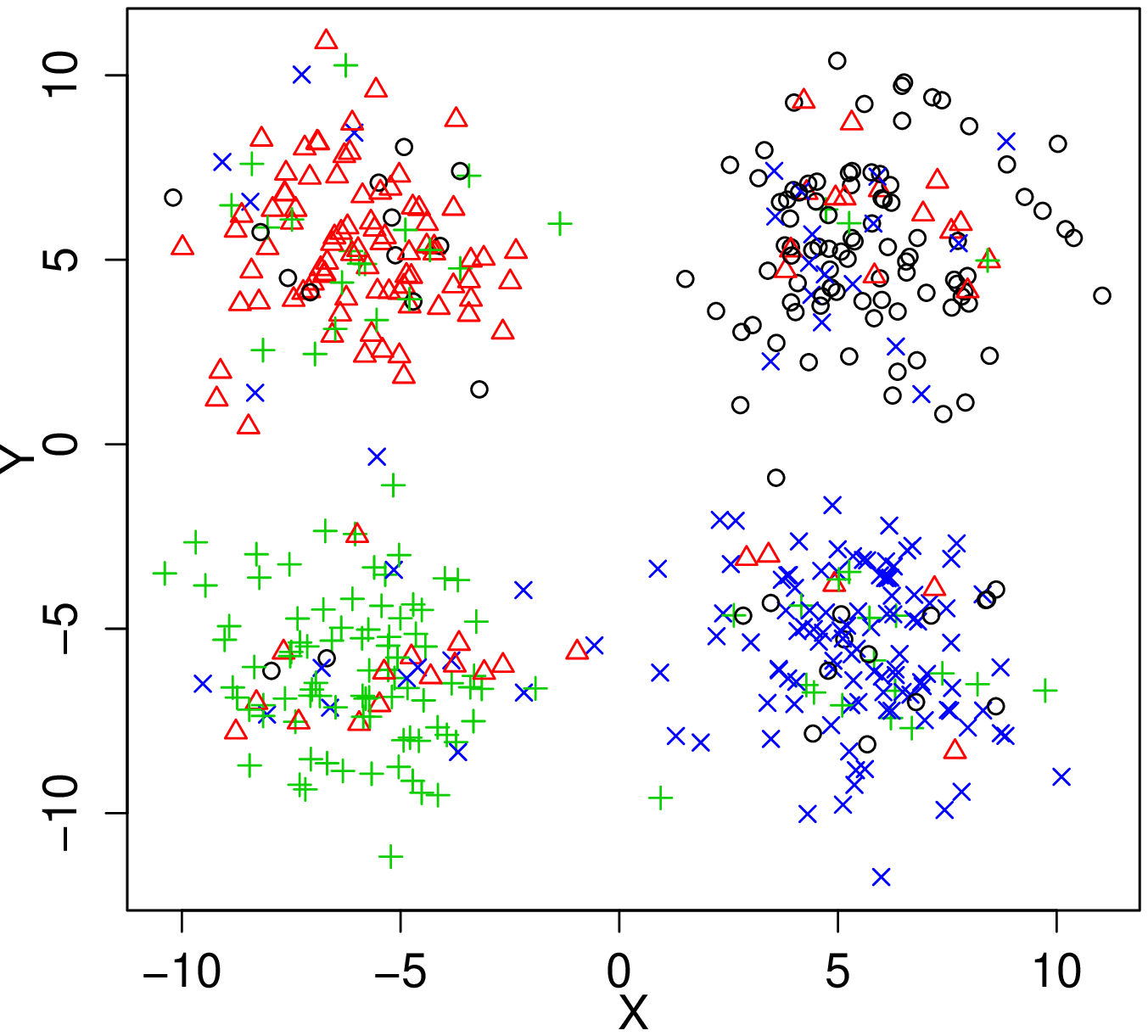}}
\caption{First, middle and last time points for the synthesis data, which contains 500 items divided into four distinct clusters. The data set is repeated for 20 different time points, showing how a random number of items drifted from one cluster to another.}
\label{fig:synthesisData}

\end{figure*}

\section{Methodology}

To measure changes in the items' behaviour or drift from their original status there should be at least more than one reading for the same attributes under consideration at different time points. The data should be separated into multiple segments based on the number of consequent readings of items' attributes so that each segment of data contains a single value of these readings. The items of each time point should be clustered using same clustering algorithm and identical parameters (like number of clusters).

After clustering, each item in every time point will be a member of a single cluster. For consequent time points items might belong to the same clusters or they might drift from one cluster into another. We might be able to quantify the amount of drifts (change of label) between any two time points by using external cluster validity indices.

As explained previously, functions of external cluster validity measures use two inputs for their calculations. The first is ground truth labels provided by experts T and second cluster groups C which are produced using one of the clustering methods, e.g. k-means. In this work we use these validity measures with different inputs. Clusters of the first time point is used instead of items' ground truth labels T to find the amount of difference between items' initial labels and their possible consequent drift from that label. Another possible method is to use consequent time points to measure changes between every pair. That is, for each time point t = {1, 2, ..., N-1} we can use any pairs of t, t+1 as inputs to the validity measure and quantify their differences.

\section{Tests and Results}
We used two types of data to measure changes over time. First, synthesis data is used to check the validity of the method and to show the efficiency of different proposed metrics to evaluate the amount of change that is happen to the data set between two time points. Second type is two different datasets for public goods game which they are used to measure the drift of players' behaviour from their initial round and other later rounds.

\subsection{Tests with Synthesised Data}
For this test a set of 500 items are generated\footnote{Data-generator and change-measure codes are available at https://goo.gl/8DBuII} on a Cartesian plane with 4 distinct clusters. The centre of these clusters is (0, 0) on the x, y plane. To simulate next time point a specific number of the items are jumped from one cluster to another by changing items' x and/or y coordinates sign. The number of items that jumped from one cluster to another is determined by a suede random function, with values between 0 and 20. Moreover a small amount of randomized change from there positions are introduced to the items to simulate the change which happened to the items within same cluster. By repeating these jumps and jiggles 19 extra time points are generated from the original dataset which all results in 20 time points. The first, middle and last time points of the dataset are shown in Fig. \ref{fig:synthesisData}

\begin{figure*}[ht]
\centering
\subfigure[First round]{%
\includegraphics[width=0.3\textwidth]{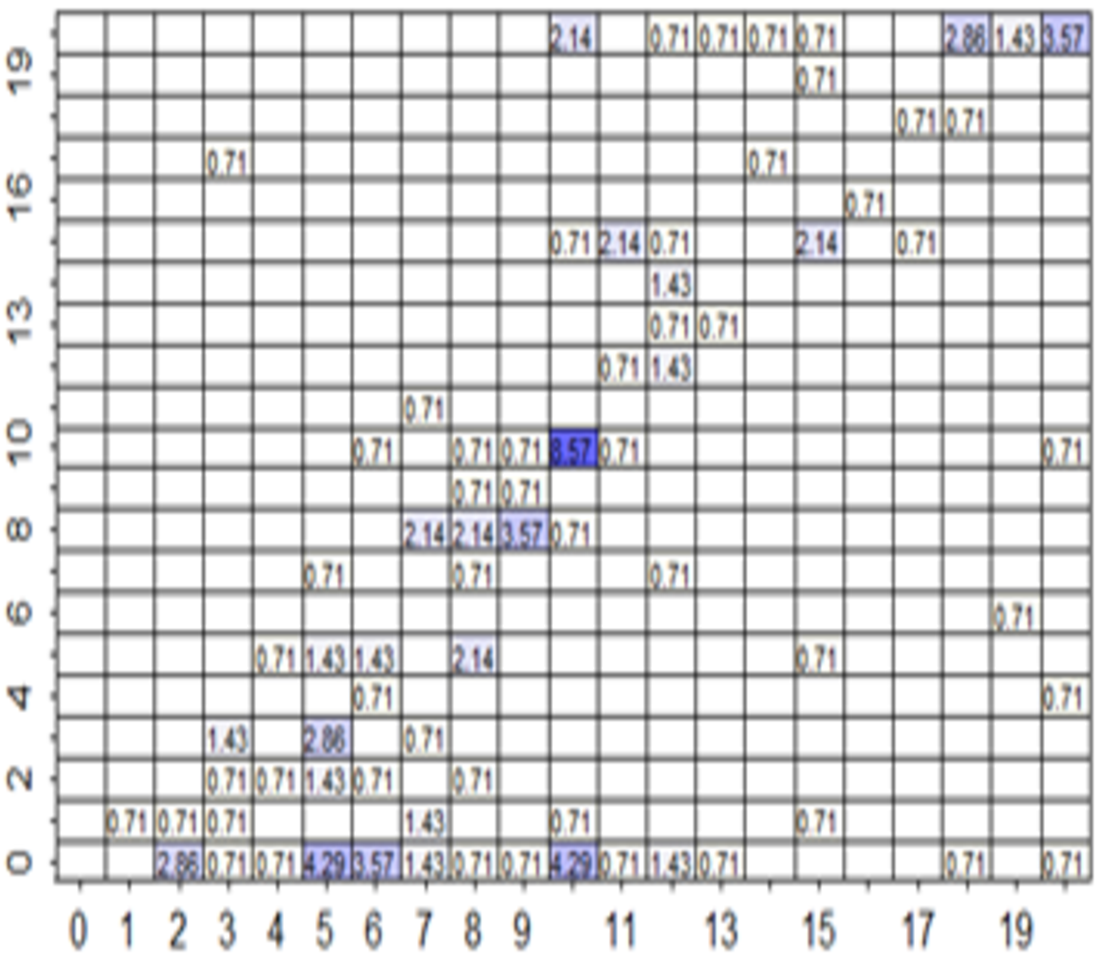}}
\subfigure[Forth round]{%
\includegraphics[width=0.3\textwidth]{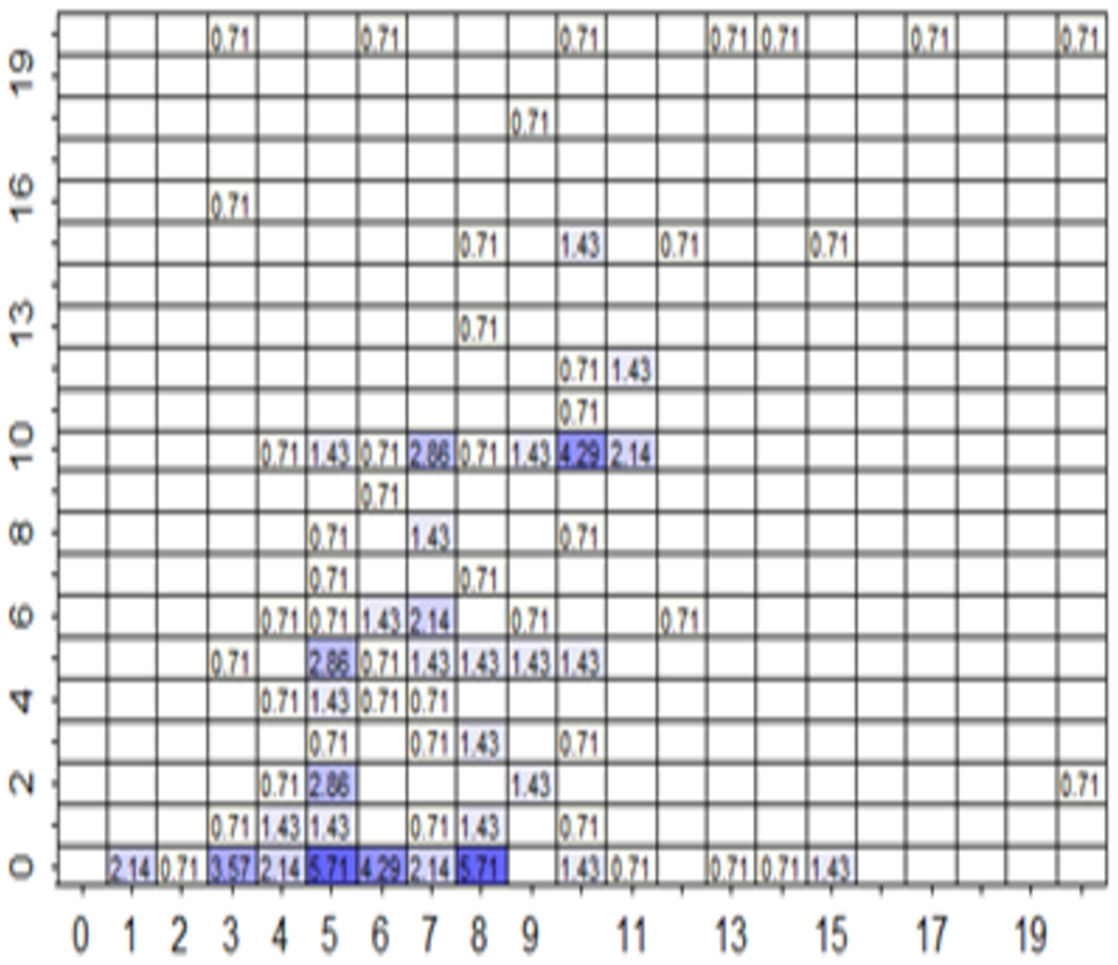}}
\subfigure[Last round]{%
\includegraphics[width=0.3\textwidth]{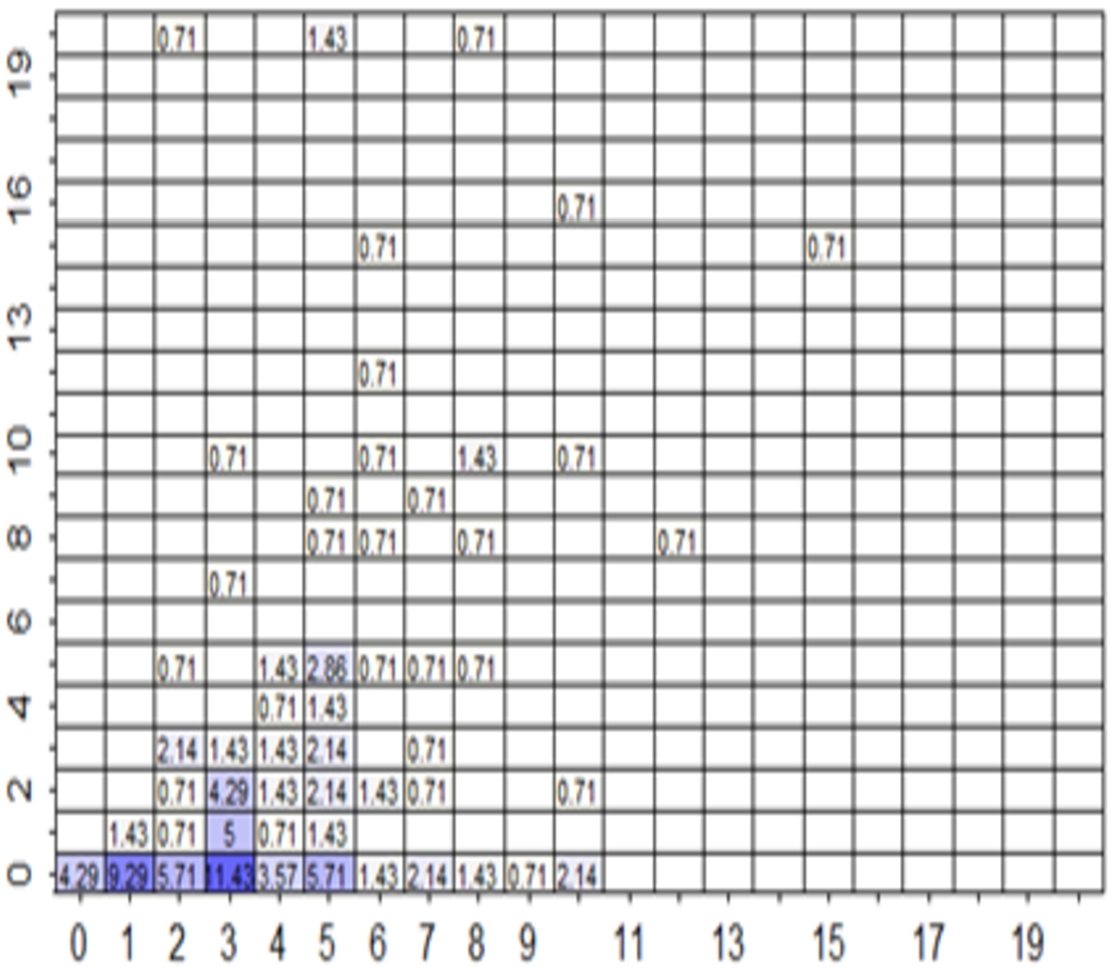}}
\caption{Player's own contribution compared to co-players' contribution in  first, fourth and last rounds of the game. }
\label{fig:pggRoundsData}

\end{figure*}

Datasets of all time points are clustered separately using k-means clustering with four clusters, and then the proposed metrics for measuring the quantity of the drift are applied on the clusters of the data by comparing the original first time point with all other time points to measure the changes which are happen in between time points. 

\begin{figure}[h]
\centering
\includegraphics[width=0.45\textwidth]{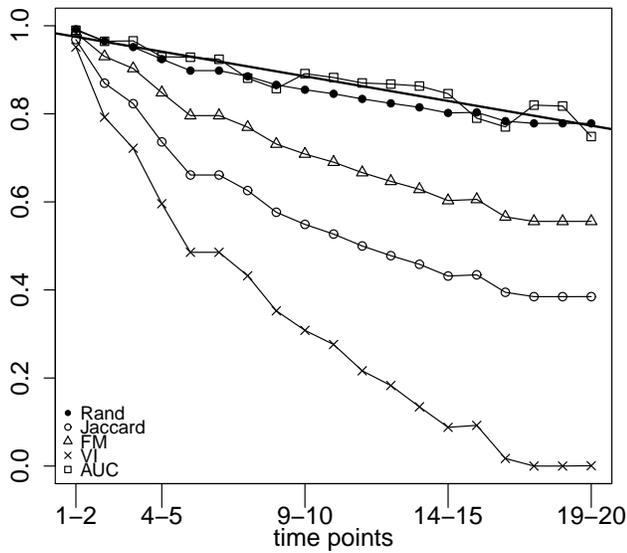}
\caption{Results of amount of shifted items between first and other time points using external cluster validity and AUC of ROC functions consequent time points.}
\label{fig:synthesisDataResult}
\end{figure}

All methods of cluster validity measures and AUC are positively recognised and quantified changes between the first time point and all other consequent time points as can be seen in Fig. \ref{fig:synthesisDataResult}. However the IV measure could not fit with other measures, as it returns zero for identical clusters, and its value increases as differences between clusters are bigger, while all other measures return one for a perfect match between clusters of two different time points, and decrease their value as difference between clusters increases. Thus VI values are scaled down and reversed using the equation $newVIResult = 1 -  \dfrac{VIresult}{ max(VIresult)}$ to have the same behaviour as other measures.

One thing to notice is that the metrics Jaccard, FM and reversed VI are exaggerating the results of the consequent changes of items' jumps, this might be due to the underlying equations as they are designed to aggressively detect miss-clustered items compared with the labels, but this might give a false perception for a grate or sudden change if used to measure differences between two time points.

\subsection{Tests with Public Goods Games Data}
Two different sets with identical attributes but different players and different time point lengths are used to measure how much players drift from their first time point readings [16]. Both data sets have attributes to identify players, determine time points, show players' own contribution, their belief about other players' contribution and other players' actual contribution in each time point. The first data set consists of 140 players, each of whom played 10 rounds, and for the second data set 128 players, each of whom played 27 rounds.

In both datasets while approaching the end of the game, in later rounds, players tend to contribute less and their expectation about other players' contribution correspondingly decreases. This is shown in Fig. \ref{fig:pggRoundsData} for the 10 rounds game data set. This drop in contribution might not be an indication of concept drift, as the players in the same class might all start to follow same pattern, hence they will remain in the same class together, which makes overall class contribution drop. 

The original players' classes provided by economists \cite{Hoice2008} are not used as their classification despite being useful to indicate the number of classes, but as mentioned before they are not based on the temporal attributes used in this experiment, so we did not depend on these labels. 

 \begin{figure}[h]
\centering
\includegraphics[width=0.45\textwidth]{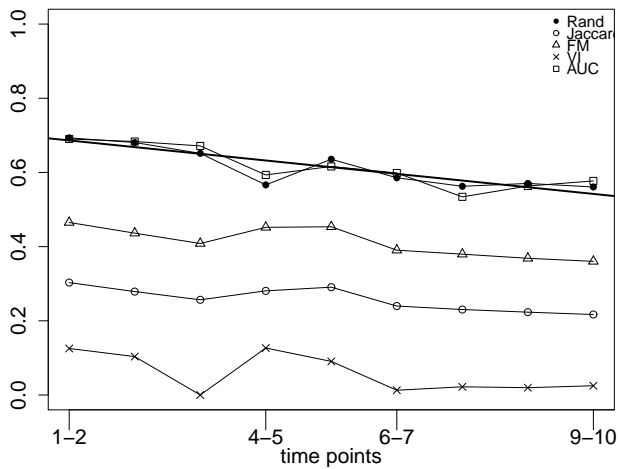}
\caption{Players' drift ratio of public goods games; 10 rounds' data comparing first and other consequent time points using Rand VI and AUC measures.}
\label{fig:game10Metrics}
\end{figure}

\begin{figure}[h]
\centering
\includegraphics[width=0.45\textwidth]{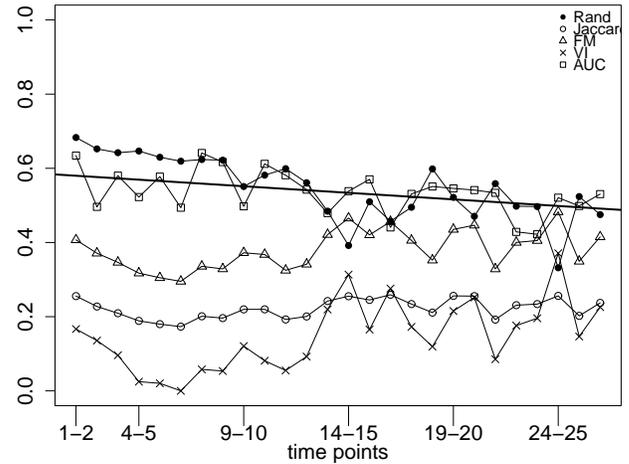}
\caption{Players' drift ratio of public goods games; 27 rounds' data comparing first and other consequent time points using Rand VI and AUC measures.}
\label{fig:game27Metrics}
\end{figure}

To find the drift ratio of players between first and next consequent game rounds we used kmeans to cluster players in each time point. Economic classifications suggest that there are four classes of players, so four clusters were used as a parameter for kmeans to cluster players in each time point, after which all measures are used to quantify players' drift from originally allocated clusters in the first round. For both data sets the results in Fig. \ref{fig:game10Metrics} and Fig. \ref{fig:game27Metrics} show negative (-0.02, -0.003 for 10 and 27 rounds) slopes of linear regression of  AUC indicating that the players tend to slowly change their behaviour toward the end of the game rounds. 

The values of AUC and Rand index of quantifying changes are comparable while other measures Jaccard, VI and FM are showing different results for the two data sets. In 10 rounds data set the later metrics are consistent with AUC and Rand as they detect more changes in the later rounds however they also show that there are more quantity of change between first and other time points. In the 27 rounds game dataset results of the later metrics (Jaccard, VI and FM) are not conclusive and they are too volatile as well as they might suggest that there is less change between first time point clusters and later consequent time points.

\section{Measuring players' Strategy Changes using MONIC}

\begin{figure}[h]
\centering
\includegraphics[width=0.45\textwidth]{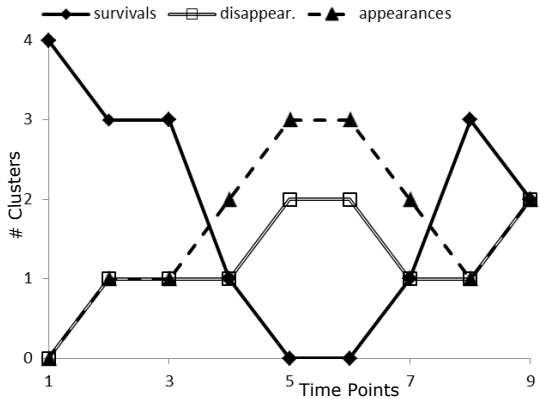}
\caption{Number of survival, appearance and disappearance of clusters between every tow consequent time points for ten rounds public goods game as measured by MONIC.}
\label{fig:monic10}
\end{figure}

\begin{figure*}[th]
\centering
\includegraphics[width=0.9\textwidth]{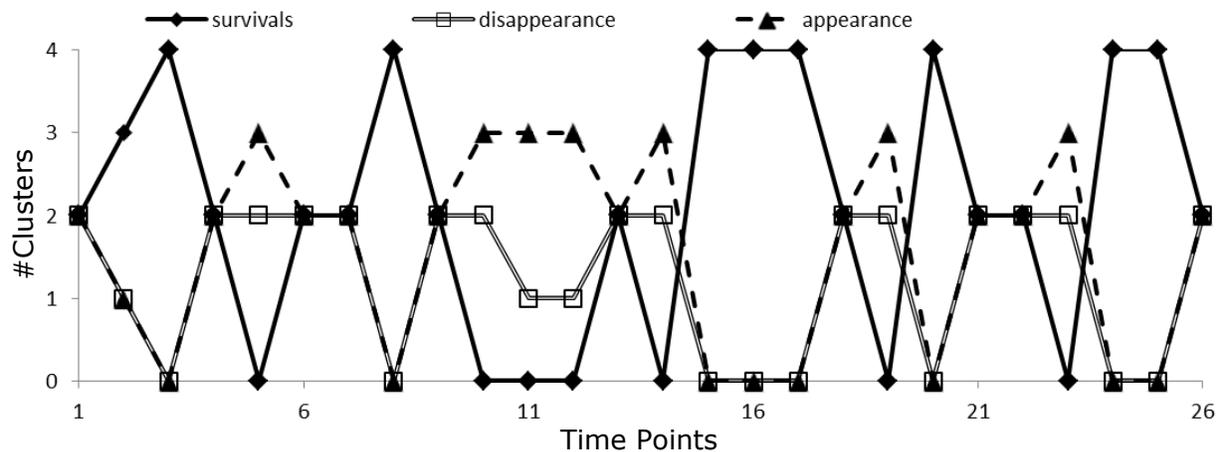}
\caption{Number of survival, appearance and disappearance of clusters between every tow consequent time points for 27 rounds public goods game as  measured by MONIC.}
\label{fig:monic27}
\end{figure*}

We used MONIC\footnote{Available at http://infolab.cs.unipi.gr/people/ntoutsi/monic.html} to  gain more insight about the public goods games data and to compare our results with the existing methods of measuring cluster changes in different time points. The data for each time period were clustered separately using k-means with four clusters. The clustering was carried out on the main temporal attributes of the data, namely belief and contribution, then the data and cluster labels of items in each consequent pair of time points was fed to the MONIC algorithm to calculate changes to clusters from one time point to another. The algorithm calculated the number of survived, appeared and disappeared clusters, as shown in figures \ref{fig:monic10} and \ref{fig:monic27}, for the ten rounds of the game.

In the first data set (10 rounds), the number of survived clusters reduced from four clusters between the first and second time points until it reached zero, while new clusters appeared in the middle of the fifth and sixth game rounds, then the number rose again until the end of the game. This might be due to the fact that players are changing their strategies and exploring new options until they ultimately settle on a certain strategic pattern. This change is consistent with our findings, as the measures slightly increase between the fifth and seventh time points, which might be an indication of players changing their strategy back to their original ones. As Keser and Winden \cite{Keser2000} suggest, this change might be due to the players responding to the average contribution of other players in the previous round.
 
The results for the 27 rounds game is not straightforward, as the numbers of cluster survivals, appearances and disappearances change more frequently. However, the cyclic pattern of increasing and decreasing number of survived clusters might be an effect of changing players' strategies or due to the underlying algorithm, as it provides an ageing factor to the items.

As the MONIC algorithm was originally introduced to detect cluster changes in data stream, it uses an ageing factor which reduces the effect of older items in the cluster and removes items older than two time points \cite{Spiliopoulou2006}. This ageing factor is essential for the algorithm to keep up-to-date with the flowing data stream and give the right results for the current status of the clusters. However, this might not be useful for public goods games data, as there is a fixed number of players which might result in the removal of players who stay in the same cluster for long time points. The effect of the ageing might not be obvious in the 10 rounds game due to the limited number of time points, but it might undermine players' strategies.

While the proposed method assumes a fixed number of clusters to calculate items membership change, the MONIC algorithm is a good method to have insights on the available clusters and their stability by measuring the number of survived clusters between two time points. However it does not measure the amount of items drifting from one cluster into another, which can be detected by the proposed method, as it introduces a specific ratio between each consequent time point, indicating the amount of change happening to the items in the clusters by calculating their membership change among clusters.

By comparing results from the proposed method and MONIC we can conclude that the players slightly and gradually change their cluster membership. The proposed method gives an exact number for the change while the MONIC presents overall clusters movement and change.

\section{Summary and Conclusions}
This paper presents a method to quantify changes over time for items across multiple readings at different times for the same attributes using external cluster validity measures and area under the curve of receiver operating characteristic. In the proposed method, each item in each time point is clustered separately with the same parameters and clustering algorithm. After clustering the amount of difference between items, cluster labels of first cluster and other consequent clusters are measured instead of ground truth labels and provided clusters.

This method proved to work on both synthetic and real public goods game data. For synthetic data measures like Rand, VI and AUC worked very well by reflecting the amount of jumps into a quantifiable scaler. Other measures like Jaccard and FM were very good at detecting changes, but they tended to exaggerate the amount of change happening in clusters' items. For the public goods game data sets, in both 10 and 27 rounds, gradually from one time point to another small numbers of players drifted from their original clusters.

Despite its useful simplicity, the proposed method has a drawback in its reliance on the first time point clusters as a reference or baseline for all other consequent time points, as the first clusters might not always be good representatives for the entire data set or latent behaviour at baseline. This might be solved by using a temporal classification method that can represent overall items' movements across all time points. However using traditional classifications algorithms might tend to exaggerate changes over multiple time points, especially when attribute values of most classes' populations change simultaneously.

\section*{Acknowledgment}
The authors record their thanks to Simon Gaechter and Felix Kolle in the School of Economics at the University of Nottingham for providing us with data from the public goods experiment and taking time to explain it to us.



\bibliographystyle{IEEEtran}
\bibliography{IEEEabrv,biblo}

\end{document}